\documentclass[11pt]{article}
\usepackage[margin=1in]{geometry}
\usepackage{times}
\usepackage{titlesec}
\usepackage{fancyhdr}
\usepackage{graphicx}
\usepackage{url}
\usepackage{authblk}
\usepackage{abstract}
\usepackage{setspace}
\usepackage{enumitem}
\usepackage{multicol}
\usepackage[numbers]{natbib}
\usepackage[hidelinks]{hyperref}

\titleformat{\section}{\large\bfseries}{\thesection.}{0.5em}{}
\titleformat{\subsection}{\normalsize\bfseries}{\thesubsection.}{0.5em}{}
\titleformat{\subsubsection}{\normalsize\itshape}{\thesubsubsection.}{0.5em}{}

\title{\textbf{Multimodal Proposal for an AI-Based Tool to Increase Cross-Assessment of Messages.}}

\author[1]{Alejandro Álvarez Castro}
\author[2]{Joaquín Ordieres-Meré}
\affil[1]{Master student at the AI master. Universidad Politécnica de Madrid. Madrid, 28040.\\ alejandro.alvarez@alumnos.upm.es}
\affil[2]{Industrial Engineering School. Universidad Politécnica de Madrid. Madrid, 28006.\\
j.ordieres@upm.es}

\date{}

\begin{document}
\maketitle

\begin{abstract}
Earnings calls represent a uniquely rich and semistructured source of financial communication, blending scripted managerial commentary with unscripted analyst dialogue. 
Although recent advances in financial sentiment analysis have integrated multimodal signals, such as textual content and vocal tone, most systems rely on flat document-level or sentence-level models, failing to capture the layered discourse structure of these interactions. 
This paper introduces a novel multimodal framework designed to generate semantically rich and structurally aware embeddings of earnings calls, by encoding them as hierarchical discourse trees. 
Each node, comprising either a monologue or a question–answer pair, is enriched with emotional signals derived from text, audio, and video, as well as structured metadata including coherence scores, topic labels, and answer coverage assessments. 
A two-stage transformer architecture is proposed: the first encodes multimodal content and discourse metadata at the node level using contrastive learning, while the second synthesizes a global embedding for the entire conference. 
Experimental results reveal that the resulting embeddings form stable, semantically meaningful representations that reflect affective tone, structural logic, and thematic alignment. 
Beyond financial reporting, the proposed system generalizes to other high-stakes unscripted communicative domains such as telemedicine, education, and political discourse, offering a robust and explainable approach to multimodal discourse representation.
This approach offers practical utility for downstream tasks such as financial forecasting and discourse evaluation, while also providing a generalizable method applicable to other domains involving high-stakes communication.
\end{abstract}


\section{Introduction}
Earnings calls are the main channels of communication through which publicly traded companies report financial results, operational highlights, and strategic directions to investors, analysts, and other parties with interest. 
Earnings calls are particularly valuable sources of financial sentiment data sources due to their structured, yet unscripted nature~\cite{Baik2024VocalCalls}. 
Quarterly sessions typically involve scripted remarks from corporate managers followed by spontaneous questions and answers sessions with financial analysts.
\citet{Todd2024Text-basedDirections} noted that earnings calls produce dense multimodal data, including verbal content, voice characteristics, and interactive dynamics between executives and analysts. 
Verbal data include explicit financial disclosures, management assessments of business performance, forward-looking guidance, and responses to analyst inquiries. 
\citet{Hajek2023SpeechPrediction} noted that earnings transcripts allow for the extraction of strategic insight through a variety of analysis techniques, including human expert analysis, AI-based search, and AI-generative summarization techniques.

Financial sentiment analysis has undergone significant methodological evolution in the past decade, transitioning from traditional lexicon-based approaches to advanced machine learning
techniques. 
The early analysis of financial sentiment relied heavily on domain-specific dictionaries and word lists that classified financial terms according to their positive or negative connotations~\citep{Nasiopoulos2025FinancialModels}.

Although recent advances in financial sentiment analysis technology have had great success integrating multimodal data, namely text and voice, to improve predictive modeling, most of these methods are sentence-level or flat document-level categorization. 
This has helped to yield huge improvements in the use of distress prediction or earnings sentiment scoring, especially by integrating pre-trained models such as FinBERT with vocal emotion features. 
However, the methodological design of the models overlooks the richer hierarchical structure of real-world financial messaging~\citep{Hajek2023SpeechPrediction,Du2024FinancialApplications}.

Earnings calls, for example, are not mere strings of speech but complex, semi-structured discussions between executives and analysts. 
These kinds of interaction change over time, according to a concrete discourse pattern consisting of monologues, direct questions, and impromptu responses, each of which involves sentences with different amounts of sentiment, coherence, and strategic intent. 
However, little in the current literature comes close to understanding earnings calls as structured discussions, and even fewer attempt to fully embed them using discourse-aware architectures~\citep{Palmieri2015ArgumentationChallenges}.

Then, research questions can be formulated accordingly,

\begin{itemize}
    \item[\textbf{RQ1:}] How can hierarchical multimodal embeddings derived from the discourse structure of earnings calls enhance the modeling of communication compared to flat unimodal representations?
    \item[\textbf{RQ2:}] To what extent do emotional trajectories, answer coherence, and interaction structure contribute to a semantically robust representation of unscripted discourse?
\end{itemize}

The overarching objective of this project is to build a unified, discourse-aware embedding of each earnings call, one that reflects both the semantic content and the emotional atmosphere conveyed throughout the interaction. These embeddings are designed to serve as compact high-level representations of the entire meeting.

In practice, such representations are intended to support downstream machine learning tasks in the financial domain, such as forecasting, predicting conference impact on the stock price, or risk assessment, especially when combined with structured financial indicators. More broadly, the system seeks to quantify the communicative tone of each conference.

Our work addresses this necessary gap by introducing a new, multimodal system that encodes the earnings call as a discourse tree. 
The process is focused on both the affective tone (from audio, text, and video modalities) and on the structural features of the conversation, such as whether or not a question was answered and how coherent the answer provided was in relation to the earlier discussion. 
By developing a two-level transformer model that operates at both the node level (dialogue pair or utterance) and the conference level (all calls), we move beyond the typical sentiment classification to a more advanced representation of communicative interactions~\citep{Todd2024Text-basedDirections}.

Specifically, such an architecture does not need domain supervision and is able to generalize to speaker roles, topics, and language. 
It is designed with self-supervised contrastive learning mechanisms to impose semantic consistency and disentangle structurally distinct interventions from each other. 
This architecture most especially makes the method adaptable to use in other areas where structured but unscripted conversations occur.

For example, in education, the same multimodal and hierarchical indicators show up in student-teacher communication, which could be addressed by real-time feedback systems that identify attentiveness. 
In governance or journalism, political discussions and media interviews involve the same blend of scripted monologue and improvisational Q\&A, quite often full of affective and rhetorical markers. 
Although in medicine, teleconsultations by physicians and patients give diagnostic insight through emotional and structural correspondence of conversation, which could be addressed using similar embeddings.

Therefore, although grounded in the financial communication context, the system presented here constitutes a more general paradigm for discourse analysis and encoding: one that captures not just what is said but also how it is said, in multiple modalities and layers of interaction. 
This situates our contribution as an initial step toward multimodal comprehension of high-stakes human conversations in a broad set of domains.

\section{Literature Review}
The growing confidence based on the coherence of the messages distributed in earnings calls is supported through several channels, including the selection of lexical terms, vocal tone, narrative structure, and rhetorical tools used by managers in presentations, as well as the question and answer sessions~\citep{Matsumoto2011WhatSessions, Fu2021TheRisk}.

Large language models (LLMs) have proved to be incredibly powerful and have gained immense attention from various fields, including banking and finance. 
Although the development of LLMs in the general domain has been extensively researched and their potential in finance is immense, research on financial LLMs (FinLLMs) is limited~\cite{Lee2024AFinLLMs}.

During the past decade, sentiment analysis in finance has evolved from basic lexicon-based tools to advanced multimodal representation learning. 
Early efforts largely focused on domain-specific word lists, such as Loughran and McDonald’s financial lexicons, which provided an initial framework to score textual sentiment. 
Although useful, these approaches lacked the nuance required to interpret tone, rhetorical strategy, and emotional cues present in natural dialogue.

The recent shift to multimodal models has brought new sophistication. 
In particular, VolTAGE~\citet{sawhney2020voltage} explored how joint modeling of text and audio from earnings calls can improve stock volatility forecasting. 
Using an intermodal attention mechanism, VolTAGE demonstrated improved performance through the interplay between verbal and vocal cues, without imposing explicit structural modeling of dialogue segments of the ACL Anthology.

Complementing this, the extensive review by~\citet{Du2024FinancialApplications} synthesized trends in financial sentiment modeling and highlighted a critical limitation: despite integrating audiovisual features, most models continue to treat entire documents or transcripts as flat inputs, overlooking discourse-level structure and hierarchical organization.

In the neuroscience-inspired domain of audio-visual sentiment processing, broad surveys underscore the importance of multimodality. 
For instance, \citet{Barua2023ADirections} identify three main fusion strategies, feature-level, decision-level, and hybrid approaches, as pillars of multimodal learning systems, applicable across text, image, and audio modalities. 
This framework supports the need for structured fusion methods capable of handling heterogeneous data.

In parallel to multimodal learning, advances in audio-focused corporate disclosure analysis have revealed the need for customized models. 
\citet{Ewertz2024ListenDisclosures} introduce FinVoc2Vec, a domain-adapted deep learning model that captures executive vocal tone in earnings calls a step forward from general emotion classifiers, which perform poorly in corporate speech. 
Their work demonstrates the predictive value of a call-specific vocal representation, especially for forecasting analyst reactions and firm performance.

Despite these advances, a gap remains: No existing financial models combine hierarchical discourse structure and multimodal fusion with self-supervised contrastive learning to deeply encode both affective nuance and interaction flow across entire earnings calls. 
VolTAGE and FinVoc2Vec underscore the value of vocal and textual data, and HMT exemplifies hierarchical transformer fusion, but none integrate these ingredients into a unified model that captures discourse units (monologue, Q\&A nodes) and global call structure.

\section{Methodology}
The main methodology approach is based on the Discourse Structure Theory (DST)~\citep{Maier2022PerspectiveModalities}. 
DST provides a cognitive and linguistic foundation for representing dialogue as a hierarchically organized structure of discourse units. 
It posits that communication is segmented into coherent units (e.g. intentions, attentional states) and governed by rhetorical relations (e.g., question-answer, elaboration, contrast).

From a theoretical perspective, the work anchors on the intersection between Multimodal Interaction Theory (MIT) and Representation Learning Theory (RLT)~\citep{Turk2014MultimodalReview,Zhang2022HypergraphMethods}.
MIT posits that communication is inherently multimodal, integrating speech, gesture, prosody, facial expression, and more. 
It emphasizes the coordinated interplay of modalities in constructing meaning in social interaction.
On the other hand, RLT underpins the use of self-supervised contrastive learning to build semantic embeddings from raw data. 
It asserts that good representations should capture essential factors of variation, be invariant to noise, and align with downstream tasks.

This work proposes a multimodal AI-based framework designed to analyze the communicative content of earnings calls, focusing on how companies convey information during these public presentations. \\
The proposed methodology consists of the following core stages and unfolds as follows.
\begin{itemize}
    \item \textit{Discourse Tree Construction}. Segment the earnings call into monologue and Q\&A nodes, encoding each with multimodal features -textual embeddings, audio emotion vectors and optional video features - along with structured metadata (topic labels, coherence scores, question-answer coverage).
    \item \textit{Node-Level Encoding}. A multimodal transformer fuses the modality features per node, enriched with metadata, to derive a fixed-dimensional embedding. Contrastive learning at this level ensures consistent representations across subsampled views within the same node.
    \item \textit{Conference-Level Encoding}. The node embeddings are sequenced in a second transformer with positional encodings and a learnable [CLS] token. Contrastive learning across perturbed node subsets encourages robustness to variations and preserves structural semantics.
    \item \textit{Training Strategy}. Hard negatives arise from different calls in the same batch, whereas positives come from multiple augmented views (utterance sampling, perturbations). Losses are back-propagated at both levels simultaneously, resulting in hierarchically aligned, semantically rich embeddings.
    \item \textit{Downstream Utility and Explainability}. The final embeddings can support downstream tasks, such as sentiment classification, financial forecasting, anomaly detection, while attention patterns, and structural metadata offer explainability.
\end{itemize}

Operationalizing this work involves several key stages: data acquisition, transcription, and metadata extraction, followed by feature identification and embedding, and finally, preparing the necessary information for model training.

\section{Implementation}
\begin{figure*}[!ht]
    \centering
    \includegraphics[width=\textwidth]{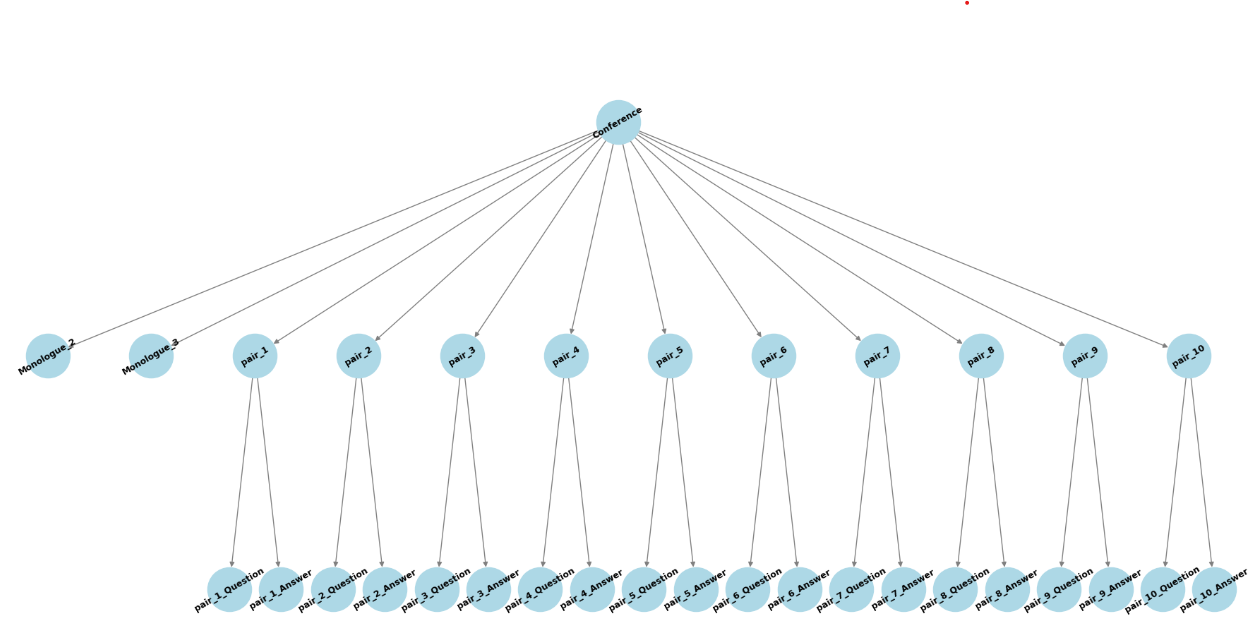}
    \caption{Tree structure representation.}
    \label{fig:conference_tree}
\end{figure*}

The details of the approach followed are presented hereinafter, where the structure of the earnings call session, including the initial presentation and further discussions is summarized in~Figure~\ref{fig:conference_tree}.

\begin{itemize}
  \item The root node labeled \textit{Conference} connects to various monologue and pair nodes and summarizes the conference.
  \item Each monologue node (e.g., \textit{Monologue\_2}) is a terminal node.
  \item Each pair node (e.g., \textit{Pair\_1}) has two children: a question node and its corresponding answer node (e.g., \textit{Pair\_1 Question} and \textit{Pair\_1 Answer}).
\end{itemize}


\subsection{Data Acquisition and Preprocessing}

The raw data consists of 3689 transcripts, each accompanied by its corresponding audio recording. Across all transcripts, a total of 260,512 individual interventions were identified and annotated with precise timestamps and metadata~\citep{Earningcalls2024EarningsCall/earningscall-python:API.}.

\begin{figure*}[!ht]
    \centering
    \includegraphics[width=\textwidth]{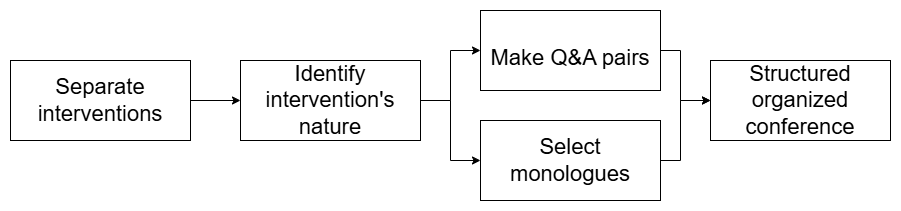}
    \caption{Preprocessing blocks diagram.}
    \label{fig:preprocessing}
\end{figure*}

As illustrated in \autoref{fig:preprocessing}, the first step in preprocessing consists of separating interventions as they appear in the raw data. These are originally segmented between the prepared remarks section—typically composed of monologues—and the remaining part corresponding to the Q\&A session.

Each intervention is then classified using large language models (LLMs) to determine its nature (e.g., Question, Answer, Monologue) and to filter out procedural or non-informative content (e.g., greetings, acknowledgments). This cleaning step ensures that only meaningful contributions are preserved. After that, questions and answers are paired by matching each detected answer to the immediately preceding detected question.

The final result is a structured and semantically coherent representation of the conference, composed exclusively of relevant and organized interventions.

In cases where the transcripts contained multilingual segments or required translation, we employed pre-trained neural machine translation (NMT) models to standardize textual inputs into English. 
NMT is a deep learning-based approach that maps text sequences from one language to another using encoder–decoder architectures, typically with recurrent or transformer-based layers. 
This step ensures consistent semantic representation across languages before emotional and structural feature extraction.

\subsection{Feature Extraction}

Each identified intervention is enriched with the following characteristics:

\begin{itemize}
\item Emotional embeddings computed at the phrase level from text, audio, and video (if available), using pretrained models. See \autoref{sec:Multi-modal Sentiment Analysis}
\item Duration of the intervention.
\item Categorical labels such as 10-K topic classification. See \autoref{sec:Extra Information}
\item Discourse-level annotations, including whether a question was answered and the coherence between responses and the initial monologue. See \autoref{sec:Extra Information}
\end{itemize}

\subsection{Embedding Construction}
Following the embedding strategy detailed later in this work, all extracted features are integrated into a single fixed length vector that represents the entire earnings call. This unified embedding captures the interaction's structure, emotional dynamics, discourse coherence, and categorical content across modalities. 

\subsection{Multi-modal Sentiment Analysis}
\label{sec:Multi-modal Sentiment Analysis}

In this section, we describe the process used to extract fine-grained emotional embeddings from different modalities (text, audio, and video). 
The analysis is performed at the sentence level rather than per intervention, in order to capture richer temporal variability and emotional nuance. 
For each sentence, the output is a 7-dimensional vector representing the intensity or probability of each of the following emotions: [anger, disgust, fear, joy, neutral, sadness, surprise].

This per-sentence granularity allows us to construct detailed emotional trajectories throughout each intervention, which are later integrated into the multimodal embedding architecture. For each modality, we employ a state-of-the-art model specialized in emotion recognition, ensuring robust and high-quality representations.

\begin{itemize}
    \item \textbf{Audio:} Sentence-level emotional embeddings are obtained using models from the \textit{Emotion2Vec} family \cite{ma2023emotion2vec}, specifically designed for speech emotion recognition.  We employ the \texttt{emotion2vec\allowbreak\_plus\allowbreak\_large} variant, which has demonstrated superior performance over alternatives like wav2vec2 or HuBERT. The original implementation is available at \href{https://github.com/ddlBoJack/emotion2vec}{this link}.

    \item \textbf{Text:} For the textual modality, we extract sentence-level emotional embeddings using the \texttt{emotion\allowbreak-english\allowbreak-distilroberta\allowbreak-base} model \cite{hartmann2022emotionenglish}. This model is based on DistilRoBERTa and fine-tuned for multiclass emotion classification across seven categories. The model offers a strong balance between performance and efficiency and is available at \href{https://huggingface.co/j-hartmann/emotion-english-distilroberta-base}{this link}.

    \item \textbf{Video:} Facial expressions are analyzed using the \texttt{vit-face-expression} model, a Vision Transformer fine-tuned for facial emotion recognition \cite{wu2020visual}. The model processes individual video frames. These frame-level predictions are then aggregated at the sentence level, using strategies such as mean, mode, or maximum pooling, to obtain a single representative emotion vector per sentence. Using the ViT architecture allows the model to capture fine-grained spatial features in facial expressions, achieving robust performance in affective computing tasks. The fine-tuned version used in our work is available at \href{https://huggingface.co/trpakov/vit-face-expression}{this other link}.
\end{itemize}

\subsection{Extra Information}
\label{sec:Extra Information}

In addition to multimodal sentiment signals, each intervention is enriched with structured analytical metadata that capture key discourse properties relevant to financial communication. This includes:

\begin{itemize}
    \item \textbf{Topic classification:} Questions and answers are mapped to regulatory categories defined in the SEC 10-K form (e.g., Risk Factors, MD\&A), providing contextual grounding for the discussion. 
    
    \item \textbf{Answer coverage:} A binary or ternary label (e.g., \textit{yes}, \textit{no}, \textit{partially}) is assigned to each question–answer pair, indicating whether the response directly addresses the question posed.
    
    \item \textbf{Discourse coherence:} For each response, coherence is evaluated relative to the surrounding monologue(s), identifying topic misalignments or logical inconsistencies.
\end{itemize}

To extract this information, we leverage a set of open-source Large Language Models (LLMs) deployed locally via \href{https://ollama.com}{Ollama}, ensuring privacy. Multiple models (e.g., \texttt{qwen2}, \texttt{gemma}, \texttt{phi}, \texttt{llama3}) are used in parallel for each task.

To improve robustness and handle conflicting or uncertain predictions, we implement a dual-layered uncertainty estimation mechanism.
\begin{itemize}
    \item \textbf{Intrinsic uncertainty} is measured by making multiple predictions with the same model and quantifying the consistency of the confidence between them.
    \item \textbf{Extrinsic uncertainty} is estimated by comparing the predictions between different models and computing the agreement between models.
\end{itemize}

The final label for each task is determined through an ensemble strategy that combines these predictions, resulting in a consensus label (e.g., whether the question was fully answered) and an aggregated confidence score. This multimodel, uncertainty-aware framework enhances both the accuracy and interpretability of the extracted information.

\subsection{Proposed method for embeddings construction}

\subsubsection{Conference Tree Organization}

We represent each earnings call as a hierarchical tree that captures the logical structure of the discourse. 
The root node encapsulates the entire conference, and two types of elements branch from it: (i) monologues, which are individual interventions without a direct reply and modeled as terminal nodes; and (ii) dialogue blocks, composed of a question and its corresponding answer, modeled as intermediate nodes with two children.

Each node integrates multimodal information, including text, audio, and optionally video embeddings, extracted at the utterance level. 
In addition, the nodes are enriched with structured metadata, such as topic labels (e.g., SEC 10-K sections), discourse coherence scores, and answer coverage assessments. This joint representation captures both the semantic content and the communicative role of each intervention within the broader context of the conference. On \autoref{fig:conference_tree} there can be seen a visualization.

\subsubsection{First Transformer (Node level)}

To produce a fixed-size embedding for each node, whether a monologue, question, or answer, we employ a Transformer-based encoder that fuses multimodal content and metadata.

The proposed model follows a dual-branch architecture. The first branch processes the sequence of utterance-level embeddings using a Transformer encoder. 
Each modality-specific embedding is first linearly projected into a shared latent space. 
A special [CLS] token is added to the sequence and multi-head self-attention is applied. The resulting output vector corresponding to the [CLS] token is extracted as the content representation of the node.

\begin{figure}[!ht]
    \centering
    \includegraphics[width=\textwidth]{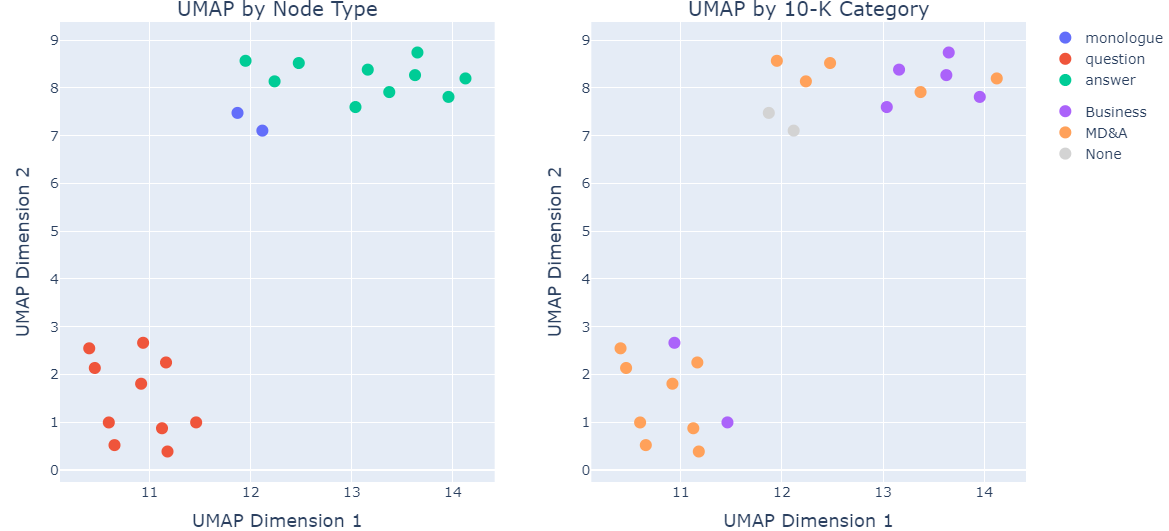}
    \caption{UMAP Projection for node-level embeddings.}
    \label{fig:node_encoder_embeddings}
\end{figure}

In parallel, the second branch encodes structured metadata, such as 10-K section labels, coherence scores, or response coverage, through a separate projection into the same latent space. The outputs of both branches are then concatenated and passed through a final linear transformation to produce the node-level embedding. In \autoref{fig:node_encoder_embeddings} we can visualize the embeddings at the node level for the conference in \autoref{fig:conference_tree}.

Given the lack of explicit supervision at the node level, we adopt a self-supervised training strategy based on contrastive learning. For each node, two different "views" are generated via subsampling of its utterances. These two views are independently encoded, resulting in two embeddings that are expected to be similar. We apply the NT-Xent loss (Normalized Temperature-scaled Cross Entropy Loss) \cite{chen2020simpleframeworkcontrastivelearning}, which encourages alignment between embeddings from the same node while pushing apart embeddings from different nodes in the same batch. This approach promotes the learning of stable and invariant representations that remain robust to internal variations while ensuring discriminability between distinct interventions.

This node-level encoder serves as the foundational module for a subsequent Transformer model that will summarize all node embeddings into a single global representation for the entire conference.

\subsubsection{Second Transformer (Conference level)}

Once each individual node in the hierarchical tree is encoded as a fixed-size embedding vector, integrating multimodal content and structured metadata, the next step is to derive a global representation for the entire conference. To achieve this, we introduce a second Transformer encoder whose goal is to aggregate the sequence of node-level embeddings into a single, semantically meaningful vector that captures the overall dynamics, structure, and communicative content of the earnings call.

The architecture follows a standard self-attention mechanism. First, a learnable token [CLS] is inserted in the sequence of node embeddings. 
Additionally, learnable positional encodings are added to the input sequence to preserve the relative order of the interventions at the conference. 
The sequence is then passed through a Transformer encoder layer consisting of multihead self-attention, residual connections, layer normalization, and a feedforward network. 
The final representation of the [CLS] token, after attention and projection, is extracted as the embedding of the full conference.

\begin{figure}[!ht]
\centering
\includegraphics[width=\textwidth]{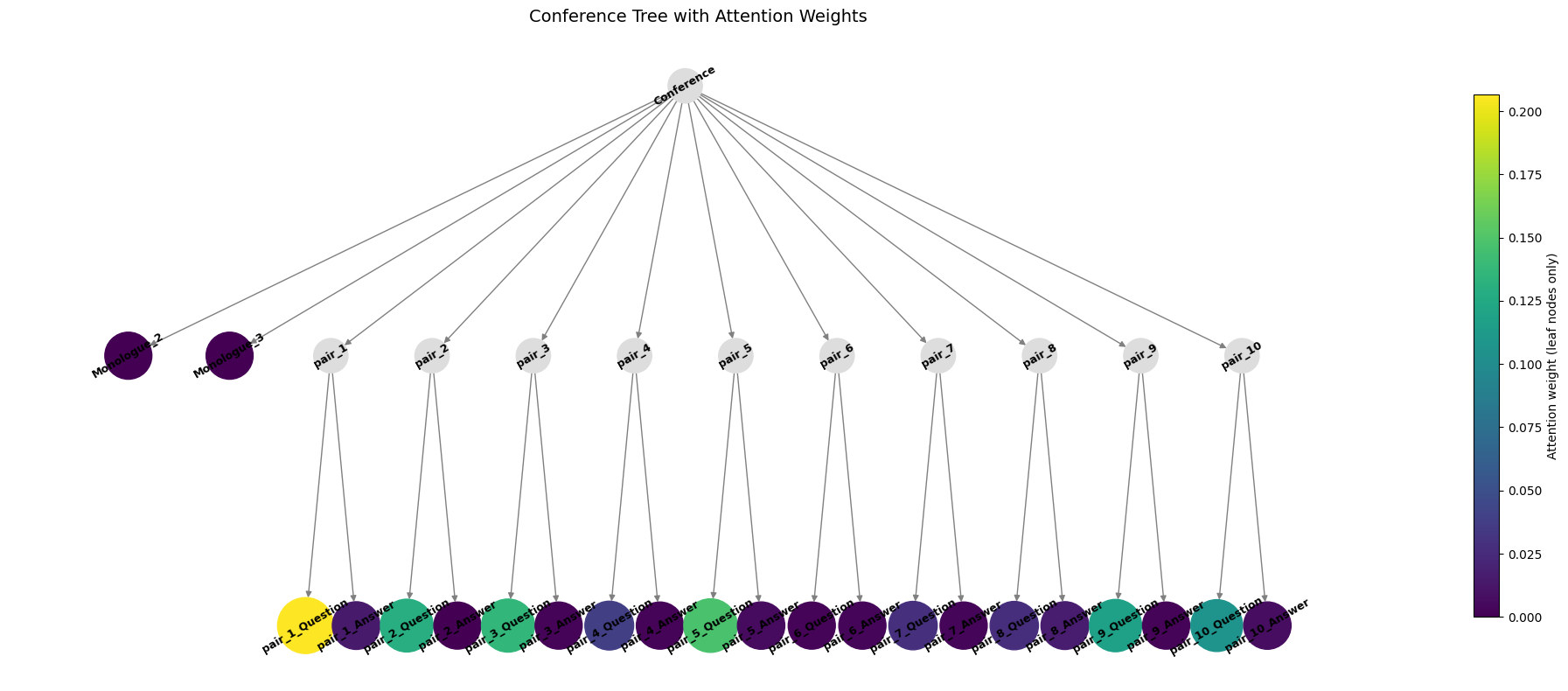}
\caption{Attention by node}
\label{fig:AttentionTree}
\end{figure}

As with node encoding, training follows a contrastive learning strategy. For each conference, two perturbed versions are generated by sampling different subsets of nodes. The encoder maps both to embeddings that are encouraged to be close in latent space via NT-Xent loss, while embeddings from different conferences are pushed apart. This promotes semantic robustness and discriminability in earnings calls.

The weights of attention of the token [CLS] toward each node also offer interpretability, as can be seen in \autoref{fig:AttentionTree}, revealing which interventions contribute the most to global meaning, an insight valuable for downstream applications.


\section{Results}
The proposed pipeline successfully generates dense, fixed-size vector representations for entire financial earnings calls, leveraging a two-stage Transformer architecture that operates hierarchically at both the node and conference levels. This architecture allows for the integration of multimodal information and structural metadata into a unified framework that summarizes the communicative and semantic content of each event.

\begin{figure}[!ht]
\centering
\includegraphics[width=\textwidth]{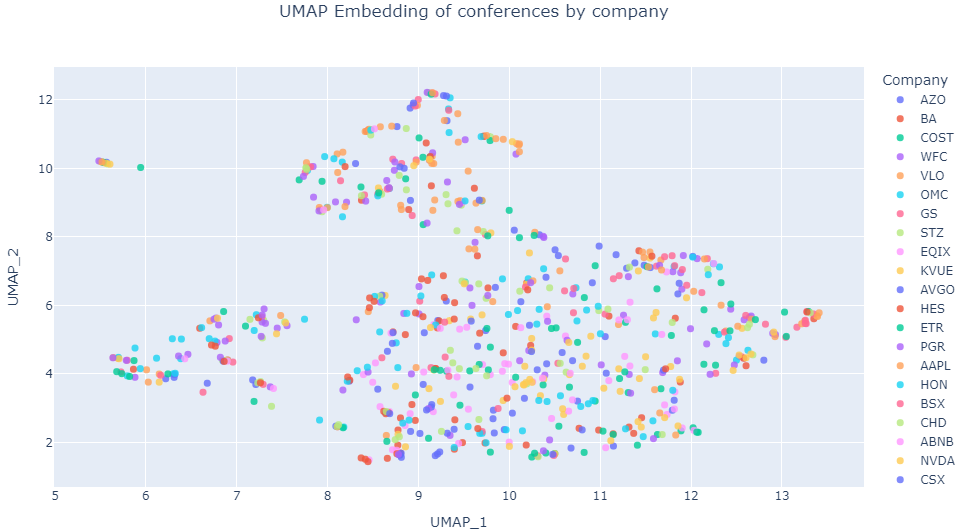}
\caption{Final embeddings}
\label{fig:Final Embeddings}
\end{figure}

To qualitatively evaluate the effectiveness of the proposed method, we project the resulting conference-level embeddings into two dimensions using UMAP, employing cosine distance as a similarity metric. This projection can be seen on \autoref{fig:Final Embeddings}.

In particular, embeddings corresponding to events with positive sentiment, strong performance announcements, or coherent executive narratives tend to be placed closer to the embedding space. 
In contrast, calls marked by uncertainty, weak performance signals, or scattered dialogue patterns appear more distant. 
In some cases, the proximity between embeddings is influenced by recurring thematic elements, such as regulatory concerns, product strategies, or market outlooks, suggesting that the attention-based encoding process captures both the topical content and the communicative structure.

To illustrate this in greater depth, we examined a pair of Tesla earnings calls from Q3 2021 and Q2 2024, represented in the UMAP projection as two distinct points situated between coordinates 5 and 6 on the x axis and between 10 and 12 on the y axis. 
Despite the temporal gap between these events, their audio embeddings demonstrated an exceptionally high cosine similarity of 0.988. This similarity was primarily due to the closely matched tone, rhythm, and emotional nuances captured in their audio embeddings.

Furthermore, thematic analysis of textual embeddings reinforced this alignment. Both earnings calls consistently discussed similar strategic themes, including production capacity, product innovation, competitive pressures, and market strategies. This thematic overlap further explains the proximity of their embeddings in the UMAP projection, indicating that the embedding generation pipeline effectively captures latent discourse-level similarities.

These findings provide promising evidence that the proposed embeddings are not only stable and discriminative, but also sensitive to the properties at the latent discourse level of financial events, making them potentially valuable for downstream tasks such as forecasting, anomaly detection, or qualitative trend analysis.


\section{Conclusions}
In this work, we have proposed a modular and hierarchical system to construct semantically rich embeddings of financial earnings calls. 
By representing each conference as a structured discourse tree and applying a dual-stage Transformer-based architecture, the system is capable of capturing both the local emotional and thematic content of individual interventions and the global communicative structure of the event. 
The resulting embeddings offer a compact yet expressive summary of each conference, integrating multimodal signals and metadata in a principled manner.
The proposed model addresses this gap by representing earnings calls as a discourse tree, encoding each node, whether a dialogue pair or monologue, with multimodal features (text, audio, optionally video), structural metadata, and a hierarchical transformer pipeline. 
Through self-supervised contrastive objectives, the model aligns node-level and call-level embeddings while preserving their distinctive roles in the discourse.

This framework lays the foundation for a wide range of downstream analyses. The qualitative projections of the embeddings demonstrate meaningful organization in the latent space, suggesting their potential utility in tasks such as classification, trend detection, or financial narrative analysis.

As a natural extension of this work, future research will focus on integrating these embeddings with market and company-level financial data. The objective is to investigate the relationship between the discourse dynamics captured in conference calls and subsequent market reactions, ultimately aiming to assess the predictive or explanatory power of such representations in financial forecasting contexts.

In addition, other applications such as the assessment of other multimedia-described evidences coming from social discussions, political arena, or others can also be considered under the proposed paradigm. 
Indeed, the proposed approach is going to be considered to describe complex scheduling in manufacturing industries, where alternative scheduling proposals can be compared and, in the end combination of different schedules can be assessed in terms of productivity and used to create reinforced learning mechanism able to pre-analyze suitability of specific scheduling proposal.


\section*{Acknowledgments}
The authors thank the partial support obtained from the Ministerio de Ciencia, Innovación y Universidades of Spain. Grant Ref. PID2022-137748OB-C31 funded by MCIN/AEI/10.13039/501100011033 and “ERDF A Way of Making Europe”.
In addition, J.O-M thanks the European Union for partial support through the RFCS program within the project DynReAct\_PDP (grant ID: 101112421).



%

\bibliographystyle{IEEEtranN}
\bibliography{biblio}

\begin{thebibliography}{20}
\providecommand{\natexlab}[1]{#1}
\providecommand{\url}[1]{#1}
\csname url@samestyle\endcsname
\providecommand{\newblock}{\relax}
\providecommand{\bibinfo}[2]{#2}
\providecommand{\BIBentrySTDinterwordspacing}{\spaceskip=0pt\relax}
\providecommand{\BIBentryALTinterwordstretchfactor}{4}
\providecommand{\BIBentryALTinterwordspacing}{\spaceskip=\fontdimen2\font plus
\BIBentryALTinterwordstretchfactor\fontdimen3\font minus \fontdimen4\font\relax}
\providecommand{\BIBforeignlanguage}[2]{{%
\expandafter\ifx\csname l@#1\endcsname\relax
\typeout{** WARNING: IEEEtranN.bst: No hyphenation pattern has been}%
\typeout{** loaded for the language `#1'. Using the pattern for}%
\typeout{** the default language instead.}%
\else
\language=\csname l@#1\endcsname
\fi
#2}}
\providecommand{\BIBdecl}{\relax}
\BIBdecl

\bibitem[Baik et~al.(2024)Baik, Kim, Kim, and Yoon]{Baik2024VocalCalls}
\BIBentryALTinterwordspacing
B.~Baik, A.~G. Kim, D.~S. Kim, and S.~Yoon, ``{Vocal Delivery Quality in Earnings Conference Calls},'' \emph{SSRN Electronic Journal}, 12 2024. [Online]. Available: \url{https://papers.ssrn.com/abstract=4398495}
\BIBentrySTDinterwordspacing

\bibitem[Todd et~al.(2024)Todd, Bowden, and Moshfeghi]{Todd2024Text-basedDirections}
\BIBentryALTinterwordspacing
A.~Todd, J.~Bowden, and Y.~Moshfeghi, ``{Text-based sentiment analysis in finance: Synthesising the existing literature and exploring future directions},'' \emph{Intelligent Systems in Accounting, Finance and Management}, vol.~31, no.~1, p. e1549, 3 2024. [Online]. Available: \url{https://onlinelibrary.wiley.com/doi/full/10.1002/isaf.1549 https://onlinelibrary.wiley.com/doi/abs/10.1002/isaf.1549 https://onlinelibrary.wiley.com/doi/10.1002/isaf.1549}
\BIBentrySTDinterwordspacing

\bibitem[Hajek and Munk(2023)]{Hajek2023SpeechPrediction}
\BIBentryALTinterwordspacing
P.~Hajek and M.~Munk, ``{Speech emotion recognition and text sentiment analysis for financial distress prediction},'' \emph{Neural Computing and Applications}, vol.~35, no.~29, pp. 21\,463--21\,477, 10 2023. [Online]. Available: \url{https://link.springer.com/article/10.1007/s00521-023-08470-8}
\BIBentrySTDinterwordspacing

\bibitem[Nasiopoulos et~al.(2025)Nasiopoulos, Roumeliotis, Sakas, Toudas, and Reklitis]{Nasiopoulos2025FinancialModels}
\BIBentryALTinterwordspacing
D.~K. Nasiopoulos, K.~I. Roumeliotis, D.~P. Sakas, K.~Toudas, and P.~Reklitis, ``{Financial Sentiment Analysis and Classification: A Comparative Study of Fine-Tuned Deep Learning Models},'' \emph{International Journal of Financial Studies 2025, Vol. 13, Page 75}, vol.~13, no.~2, p.~75, 5 2025. [Online]. Available: \url{https://www.mdpi.com/2227-7072/13/2/75/htm https://www.mdpi.com/2227-7072/13/2/75}
\BIBentrySTDinterwordspacing

\bibitem[Du et~al.(2024)Du, Xing, Mao, and Cambria]{Du2024FinancialApplications}
\BIBentryALTinterwordspacing
K.~Du, F.~Xing, R.~Mao, and E.~Cambria, ``{Financial Sentiment Analysis: Techniques and Applications},'' \emph{ACM Computing Surveys}, vol.~56, no.~9, p. 220, 10 2024. [Online]. Available: \url{https://dl.acm.org/doi/pdf/10.1145/3649451}
\BIBentrySTDinterwordspacing

\bibitem[Palmieri et~al.(2015)Palmieri, Rocci, and Kudrautsava]{Palmieri2015ArgumentationChallenges}
\BIBentryALTinterwordspacing
R.~Palmieri, A.~Rocci, and N.~Kudrautsava, ``{Argumentation in earnings conference calls. Corporate standpoints and analysts’ challenges},'' \emph{Studies in Communication Sciences}, vol.~15, no.~1, pp. 120--132, 1 2015. [Online]. Available: \url{https://www.sciencedirect.com/science/article/abs/pii/S1424489615000338?utm_source=chatgpt.com}
\BIBentrySTDinterwordspacing

\bibitem[Matsumoto et~al.(2011)Matsumoto, Pronk, and Roelofsen]{Matsumoto2011WhatSessions}
\BIBentryALTinterwordspacing
D.~Matsumoto, M.~Pronk, and E.~Roelofsen, ``{What Makes Conference Calls Useful? The Information Content of Managers' Presentations and Analysts' Discussion Sessions},'' \emph{The Accounting Review}, vol.~86, no.~4, pp. 1383--1414, 7 2011. [Online]. Available: \url{https://dx.doi.org/10.2308/accr-10034}
\BIBentrySTDinterwordspacing

\bibitem[Fu et~al.(2021)Fu, Wu, and Zhang]{Fu2021TheRisk}
\BIBentryALTinterwordspacing
X.~Fu, X.~Wu, and Z.~Zhang, ``{The Information Role of Earnings Conference Call Tone: Evidence from Stock Price Crash Risk},'' \emph{Journal of Business Ethics}, vol. 173, no.~3, pp. 643--660, 10 2021. [Online]. Available: \url{https://link.springer.com/article/10.1007/s10551-019-04326-1}
\BIBentrySTDinterwordspacing

\bibitem[Lee et~al.(2024)Lee, Stevens, Han, and Song]{Lee2024AFinLLMs}
\BIBentryALTinterwordspacing
J.~Lee, N.~Stevens, S.~C. Han, and M.~Song, ``{A Survey of Large Language Models in Finance (FinLLMs)},'' \emph{Neural Computing and Applications 2025}, pp. 1--15, 2 2024. [Online]. Available: \url{http://arxiv.org/abs/2402.02315}
\BIBentrySTDinterwordspacing

\bibitem[Sawhney et~al.(2020)Sawhney, Khanna, Aggarwal, Jain, Mathur, and Shah]{sawhney2020voltage}
R.~Sawhney, P.~Khanna, A.~Aggarwal, T.~Jain, P.~Mathur, and R.~Shah, ``Voltage: Volatility forecasting via text audio fusion with graph convolution networks for earnings calls,'' in \emph{Proceedings of the 2020 conference on empirical methods in natural language processing (EMNLP)}, 2020, pp. 8001--8013.

\bibitem[Barua et~al.(2023)Barua, Ahmed, and Begum]{Barua2023ADirections}
A.~Barua, M.~U. Ahmed, and S.~Begum, ``{A Systematic Literature Review on Multimodal Machine Learning: Applications, Challenges, Gaps and Future Directions},'' \emph{IEEE Access}, vol.~11, pp. 14\,804--14\,831, 2023.

\bibitem[Ewertz et~al.(2024)Ewertz, Knickrehm, Nienhaus, and Reichmann]{Ewertz2024ListenDisclosures}
\BIBentryALTinterwordspacing
J.~Ewertz, C.~Knickrehm, M.~Nienhaus, and D.~Reichmann, ``{Listen Closely: Measuring Vocal Tone in Corporate Disclosures},'' \emph{SSRN Electronic Journal}, 12 2024. [Online]. Available: \url{https://papers.ssrn.com/abstract=4307178}
\BIBentrySTDinterwordspacing

\bibitem[Maier and Steinbach(2022)]{Maier2022PerspectiveModalities}
\BIBentryALTinterwordspacing
E.~Maier and M.~Steinbach, ``{Perspective Shift Across Modalities},'' \emph{Annual Review of Linguistics}, vol.~8, no. Volume 8, 2022, pp. 59--76, 1 2022. [Online]. Available: \url{https://www.annualreviews.org/content/journals/10.1146/annurev-linguistics-031120-021042}
\BIBentrySTDinterwordspacing

\bibitem[Turk(2014)]{Turk2014MultimodalReview}
M.~Turk, ``{Multimodal interaction: A review},'' \emph{Pattern Recognition Letters}, vol.~36, no.~1, pp. 189--195, 1 2014.

\bibitem[Zhang et~al.(2022)Zhang, Guo, Wang, Li, Zhang, Li, Liu, Zhang, Guo, Wang, Wang, Li, and Zhang]{Zhang2022HypergraphMethods}
\BIBentryALTinterwordspacing
L.~. Zhang, J.~. Guo, J.~. Wang, S.~. Li, C.~Zhang, J.~Li, J.~Liu, L.~Zhang, J.~Guo, J.~Wang, J.~Wang, S.~Li, and C.~Zhang, ``{Hypergraph and Uncertain Hypergraph Representation Learning Theory and Methods},'' \emph{Mathematics 2022, Vol. 10, Page 1921}, vol.~10, no.~11, p. 1921, 6 2022. [Online]. Available: \url{https://www.mdpi.com/2227-7390/10/11/1921/htm https://www.mdpi.com/2227-7390/10/11/1921}
\BIBentrySTDinterwordspacing

\bibitem[{Earningcalls}(2024)]{Earningcalls2024EarningsCall/earningscall-python:API.}
\BIBentryALTinterwordspacing
{Earningcalls}, ``\BIBforeignlanguage{English}{{EarningsCall/earningscall-python: The EarningsCall Python library provides convenient access to the EarningsCall API.}}'' 11 2024. [Online]. Available: \url{https://github.com/EarningsCall/earningscall-python}
\BIBentrySTDinterwordspacing

\bibitem[Ma et~al.(2024)Ma, Zheng, Ye, Li, Gao, Zhang, and Chen]{ma2023emotion2vec}
Z.~Ma, Z.~Zheng, J.~Ye, J.~Li, Z.~Gao, S.~Zhang, and X.~Chen, ``emotion2vec: Self-supervised pre-training for speech emotion representation,'' \emph{Proc. ACL 2024 Findings}, 2024.

\bibitem[Hartmann(2022)]{hartmann2022emotionenglish}
J.~Hartmann, ``Emotion english distilroberta-base,'' \url{https://huggingface.co/j-hartmann/emotion-english-distilroberta-base/}, 2022.

\bibitem[Wu et~al.(2020)Wu, Xu, Dai, Wan, Zhang, Yan, Tomizuka, Gonzalez, Keutzer, and Vajda]{wu2020visual}
B.~Wu, C.~Xu, X.~Dai, A.~Wan, P.~Zhang, Z.~Yan, M.~Tomizuka, J.~Gonzalez, K.~Keutzer, and P.~Vajda, ``Visual transformers: Token-based image representation and processing for computer vision,'' 2020.

\bibitem[Chen et~al.(2020)Chen, Kornblith, Norouzi, and Hinton]{chen2020simpleframeworkcontrastivelearning}
\BIBentryALTinterwordspacing
T.~Chen, S.~Kornblith, M.~Norouzi, and G.~Hinton, ``A simple framework for contrastive learning of visual representations,'' 2020. [Online]. Available: \url{https://arxiv.org/abs/2002.05709}
\BIBentrySTDinterwordspacing

\end{thebibliography}

\section*{Authors}
\noindent
\begin{minipage}{0.7\textwidth}
\textbf{Alejandro \'Alvarez Castro} was born and raised in Madrid, Spain, where he completed a bilingual secondary education program, earning diplomas from both the Spanish and French systems. 
He holds a Bachelor's degree in Engineering and Data Systems from the Universidad Politécnica de Madrid (UPM) / Technological University of Madrid, and is currently completing an academic Master's in Big Data and Machine Learning at the same institution. 
Professionally, he has over a year and a half of experience in data governance consulting, delivering tailored, data-driven solutions across various industries. 
Fluent in Spanish, English, and French, he is driven by a strong interest in applying artificial intelligence to real-world challenges beyond the academic domain: to optimize, automate, or improve seemingly complex processes across different sectors. He is also cultivating a growing interest in the intersection between AI and finance.
\end{minipage}%
\hfill%
\begin{minipage}{0.22\textwidth}
\includegraphics[width=\linewidth]{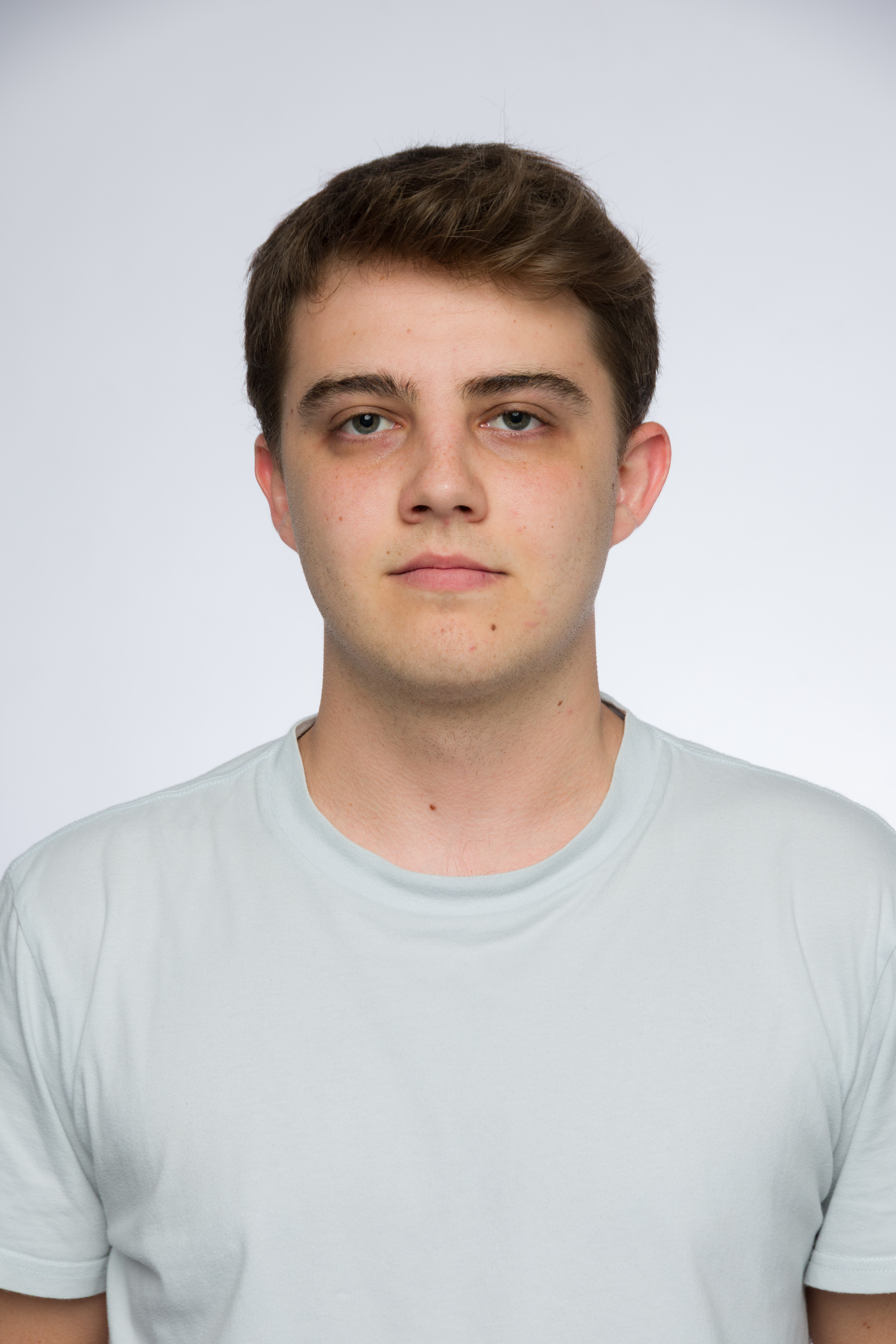}
\end{minipage}

\vspace{1em}%

\noindent
\begin{minipage}{0.7\textwidth}
\textbf{Joaqu\'in Ordieres-Mer\'e} is a Full Professor at the Universidad Polit\'ecnica de Madrid, affiliated with the Department of Industrial Engineering, Business Administration and Statistics. With a career rooted in engineering and data-driven innovation, his research spans a wide range of domains including knowledge management, big data, business intelligence, industrial process optimization, and environmental modeling. He has contributed extensively to the advancement of industry 4.0 concepts, with a particular focus on human behavior interaction, the steel industry, and the Industrial Internet of Things. Prof. Ordieres-Meré is also actively involved in consulting and applied research, leveraging data modeling techniques such as finite element methods and numerical modeling to solve complex industrial and environmental challenges.
\end{minipage}%
\hfill%
\begin{minipage}{0.22\textwidth}
\includegraphics[width=\linewidth]{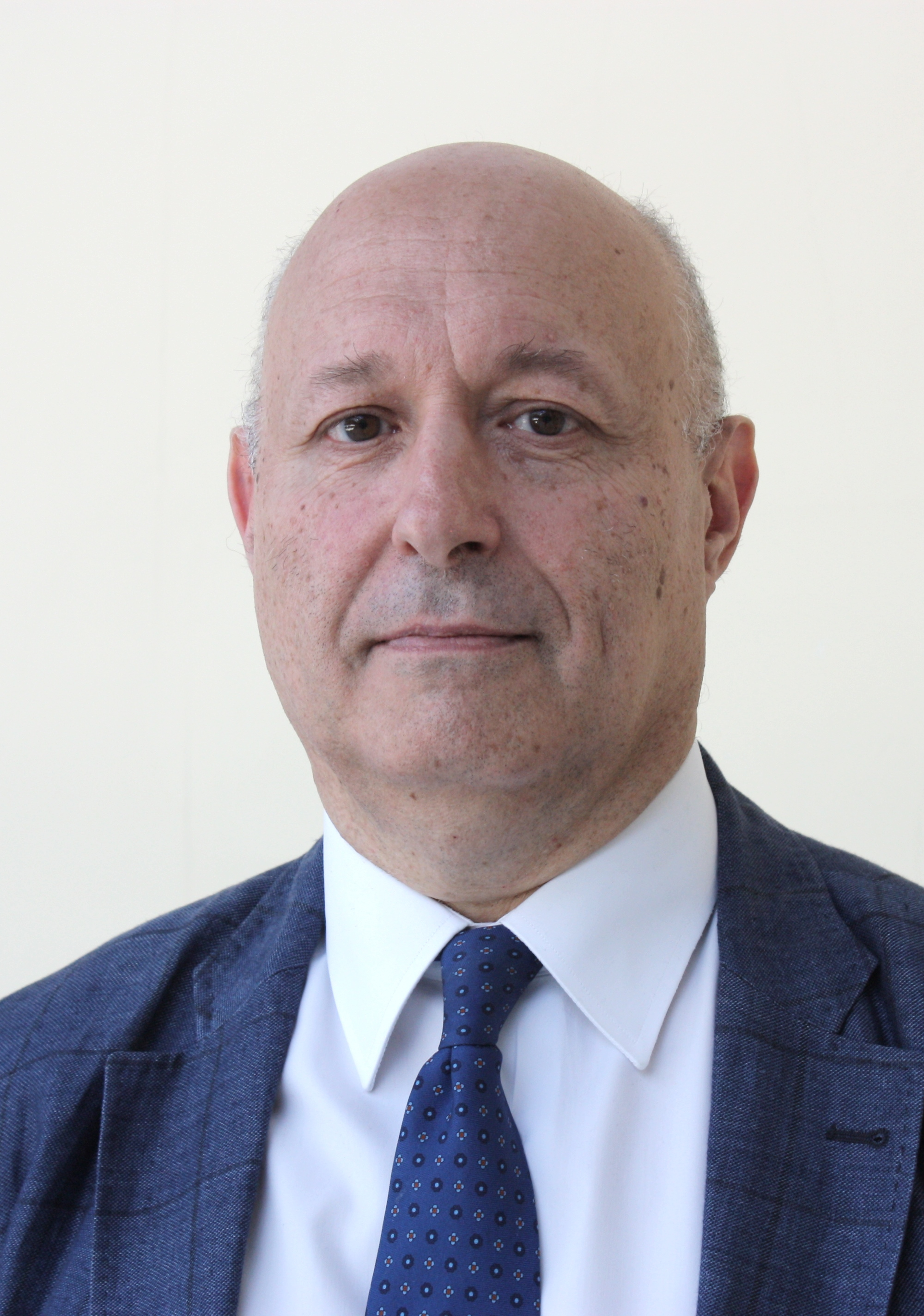}
\end{minipage}

\end{document}